\documentclass{article}
\usepackage{spconf,amsmath,graphicx}

\usepackage{times}
\usepackage{epsfig}
\usepackage{graphicx}
\usepackage{amsmath}

\usepackage{amssymb}
\usepackage{algorithm}
\usepackage{algorithmic}
\usepackage{multirow}
\usepackage{booktabs}
\usepackage{array,color}
\usepackage{caption}
\usepackage{subfigure}
\usepackage{float}
\usepackage{supertabular}
\usepackage{longtable}

\usepackage{CJKutf8}

\title{DATA AUGMENTATION IMBALANCE FOR Imbalanced Attribute Classification }
\name{{\sthanks{Yang Hu and Xiaying Bai have contributed equally to this work.}}Yang Hu$^{\star \dagger}$ \qquad Xiaying Bai$^{\star \dagger}$ \qquad  Pan Zhou$^{\star}$ \qquad Fanhua Shang $^{\star}$ \qquad Shengmei Shen $^{\dagger}$}
\address{${\star \dagger}$ huyangtorus@gmail.com  \qquad $^{\star \dagger}$ xyBai$\_$1@stu.xidian.edu.cn \qquad $^{\star}$ pzhou@u.nus.edu \\ $^{\star}$ fhshang@xidian.edu.cn \qquad $^{\dagger}$ shengmei.shen@sg.panasonic.com}

\begin{document}
%
\maketitle
\begin{abstract}
Pedestrian attribute recognition is an important multi-label classification problem. Although the convolutional neural networks are prominent in learning discriminative features from images, the data imbalance in multi-label setting for fine-grained tasks remains an open problem. In this paper, we propose a new re-sampling algorithm called: data augmentation imbalance (DAI) to explicitly enhance the ability to discriminate the fewer attributes via increasing the proportion of labels accounting for a small part. Fundamentally, by applying over-sampling and under-sampling on the multi-label dataset at the same time, the thought of robbing the rich attributes and helping the poor makes a significant contribution to DAI. Extensive empirical evidence shows that our DAI algorithm achieves state-of-the-art results, based on pedestrian attribute datasets, i.e. standard PA-100K and PETA datasets.

\end{abstract}
\begin{keywords}
Pedestrian attribute recognition, data imbalance, multi-label classificaction, sampling, deep learning
\end{keywords}
\section{Introduction}
\label{sec:intro}
Pedestrian attribute such as age, gender and clothing style are humanly searchable semantic descriptions and can be applied in visual surveillance.

Although many works have been proposed on this topic, however, pedestrian attribute recognition is still a problem due to challenging factors, such as viewpoint change, low illumination, low resolution and so on. Particularly, many data in person attribute recognition naturally exhibit imbalance in their class distribution. For example, the proportion of the attribute of lowerBodyCasual reaches 0.86, accessorNothing reaches 0.74, while the samples at the age larger 60 only take up 0.06 on PETA dataset \cite{deng2014pedestrian}. Attributes with high proportions are highly skewed since it is easier to obtain samples than those with fewer samples in real life. Such attribute recognition problems provide excellent testbeds for studying multi-label imbalanced learning algorithms.

To deal with these, deep learning has achieved an appalling performance due to their success in automatic feature extraction. Several attribute recognition algorithms based on deep learning has been proposed on these breakthroughs, such as \cite{li2015multi,  liu2018localization, liu2017hydraplus,sarfraz2017deep,zhao2018grouping,wang2017attribute}. However, they barely pay attention to the heavy imbalance of the data in PAR. Even \cite{li2015multi} uses weighted loss to mitigate the imbalance, its efficiency is still poor compared to our DAI. Data for face attribute analysis often display highly-skewed label distribution, \cite{huang2019deep} implements a deep neural network to maintain inter-cluster margins both within and between classes, but it's hard to distinguish the margin within the classes and to cluster attributes in PAR. The authors introduce a loss function \cite{sarafianos2018deep} to handle class imbalance both at class and at an instance level to retard the imbalance issue in PAR.
Yet there is still no new insights into the imbalance problem in sampling methods in multi-label setting.

\begin{figure}[!ht]
\setlength{\abovecaptionskip}{0.cm}
	\begin{center}
		\subfigure[]{
			\label{fig_1}
			\centering
			\includegraphics[width=2.5 in]{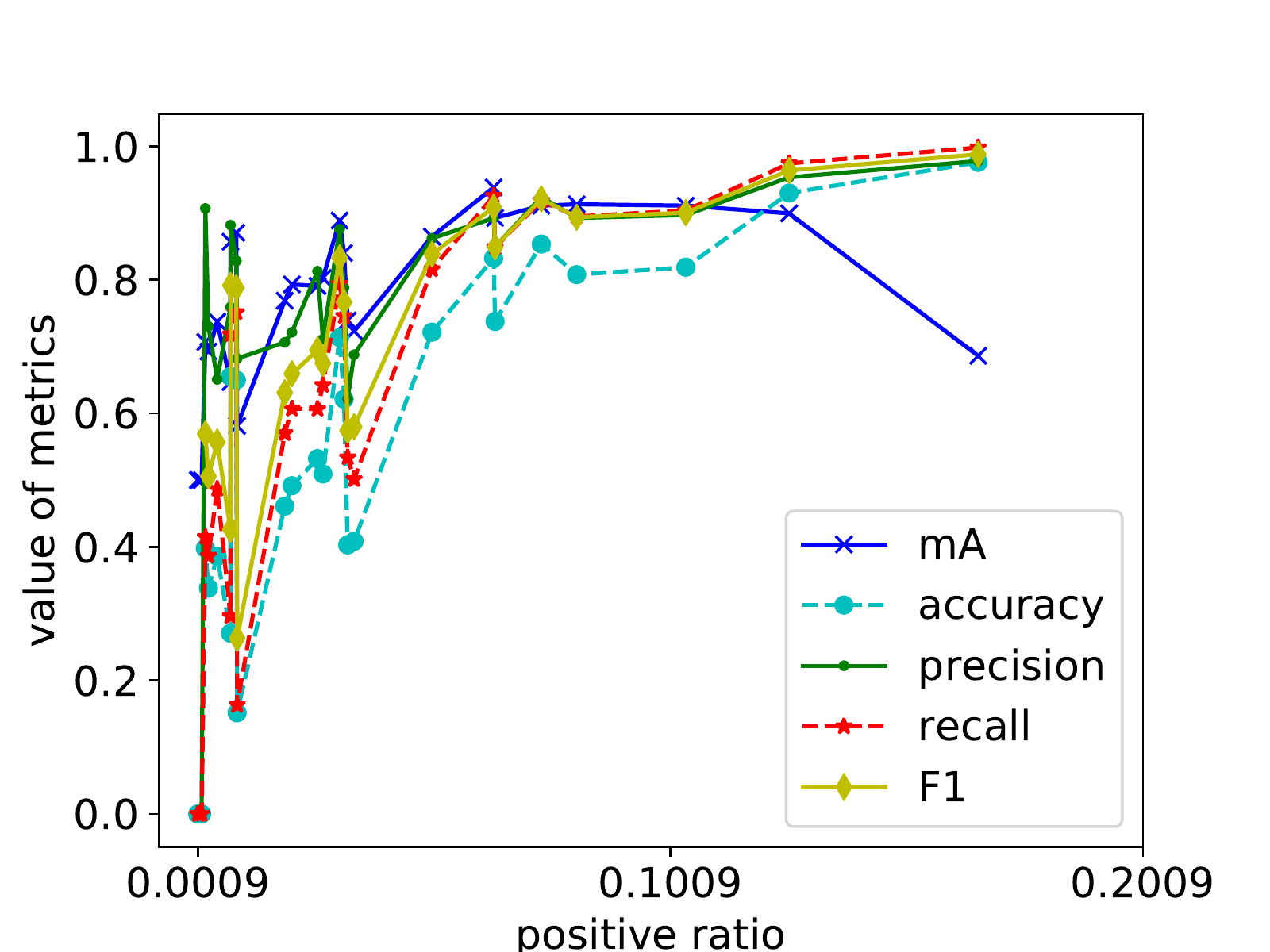}      
		}\hspace{3mm}
		\subfigure[]{
			\label{fig_2}
			\centering
			\includegraphics[width=2.5 in]{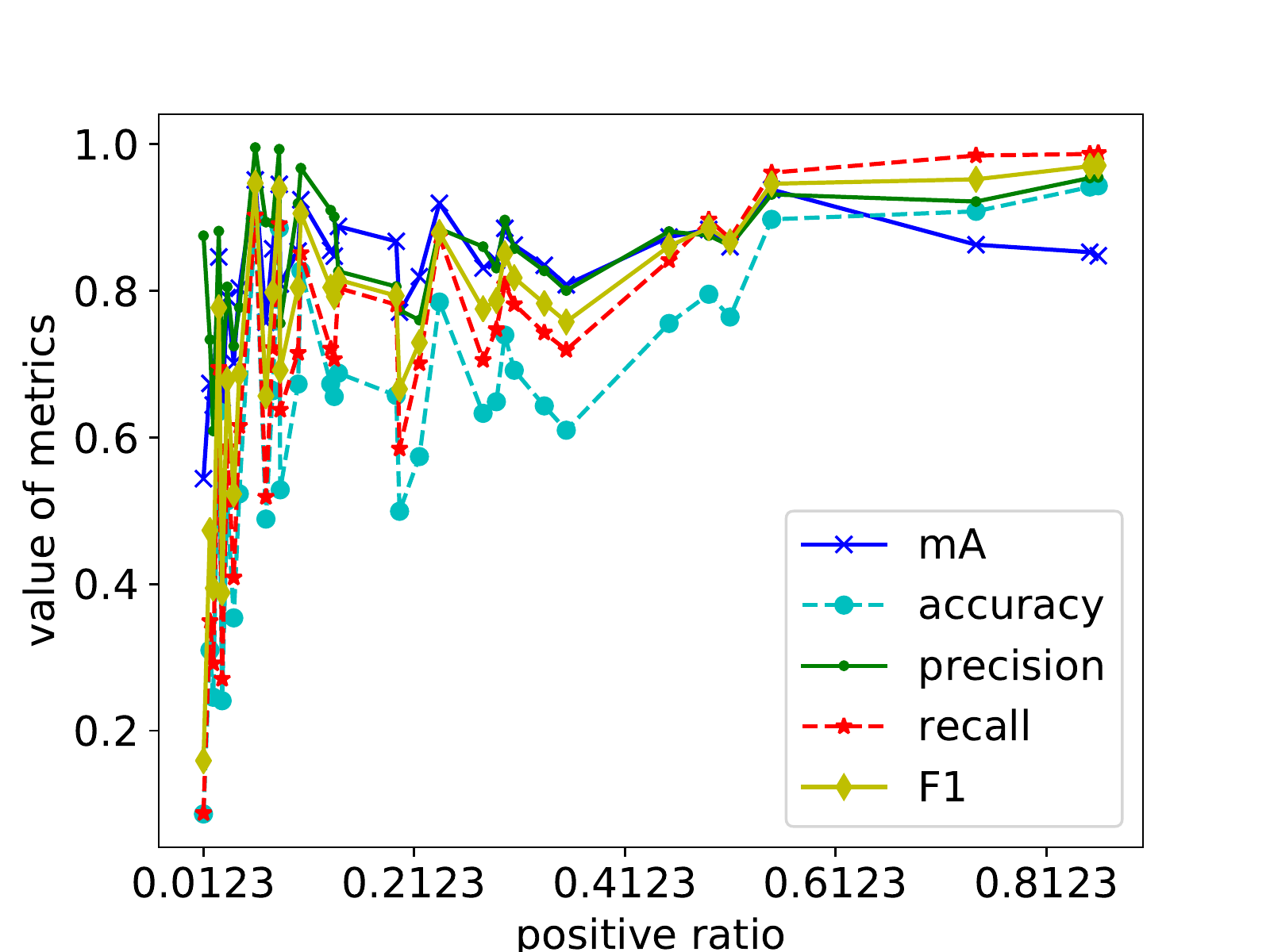}     
		}\hspace{-30mm}
	\end{center}
	\caption{Two figures depict the relation between the value of metrics and the proportion of positive labels, (a) is PA-100K dataset and (b) is PETA dataset.} \label{fig_12}
\end{figure}

Figure \ref{fig_12} shows how the imbalance of the datasets(PA-100K\cite{liu2017hydraplus}, PETA\cite{deng2014pedestrian}) harms the performance of PAR task. We tested 5 metrics including mA, accuracy, precision, recall, F1 scores with Densenet161 as the backbone. We can see that labels with a low proportion of positive samples have a low and unstable performance on the multi-label classification. To retard the negative effect of data imbalance, impressive results have been proposed for imbalanced data learning. One kind of approach called cost-sensitive learning \cite{ting2000comparative,zadrozny2003cost,zhou2005training} adjusting the decision procedure in a way is exploted. Nevertheless, it is tough to determine the exact cost for different samples in various distributions. Another kind of approach is re-sampling\cite{chawla2002smote,he2009learning,maciejewski2011local}, which either oversamples the minority class data or undersamples the majority class data to balance the distribution of data. But these sampling methods are not able to work in the multi-label setting, as the augmentation of one label may harm the balance of the other label.

To solve the issue of resampling in the multi-label scenario, DAI focuses on the imbalance data in the multi-label setting. Then we use DAI to construct the sub-balance dataset, then we train the sub-balance and the whole train set in turn.
The traits of our algorithm: 1. DAI explicitly enhance discriminative ability of fewer attributes via increasing the positive ratio of labels accounting for a small part. 2. DAI can work in the multi-label setting. It can balance each label as much as possible.
In the following extensive evaluation, we achieve the state-of-the-art results in 2 mainstream datasets(PA-100K, PETA), our method achieves significant and consistent improvement as compared to other baselines. 

\section{Related Work}

\textbf{Pedestrian Attribute Recognition\quad}Pedestrian attribute recognition is a multi-label classification problem that has been extensively applied to video surveillance systems. Previous works typically train a separate classifier (e.g. SVM or AdaBoost) for each attribute based on handcrafted features such as color histograms and local binary patterns while ignoring the relationship among pedestrian attributes \cite{deng2014pedestrian, layne2014attributes, layne2012person}. Besides, the great challenge originates from the large underlying class imbalance. To address the two drawbacks, DeepMAR \cite{li2015multi}, a deep learning framework, can recognize multiple attributes jointly with a novel weighted loss alleviating the multi-label imbalance. \cite{wang2017attribute} offers an end-to-end encoder/decoder recurrent network to learn the sequential ordering dependencies and correlations among attributes. Commonly-used practice such as attention model \cite{sarafianos2018deep} extract and aggregate visual attention masks at different scales. Particularly, a classical attention-based deep neural network is proposed in 2017, named as HydraPlus-Net \cite{liu2017hydraplus}, which explore the multi-scale attentive features to enrich the feature representations. Different from previous works, we propose a re-sampling method to improve discrimination in multi-label classification problem with jointly recognizing attributes.

\textbf{Imbalanced classification.\quad}Machine learning methods for handling class imbalance are grouped into data-level and algorithm-level methods. Data-level methods aim to alter the training data distribution to decrease imbalance and learn equally good classifiers for all classes, usually by random under-sampling and over-sampling techniques. Although informed over-sampling techniques have been developed to strengthen class boundaries, a well-known issue with the replication-based feature is its tendency to over-fit. To this end, SMOTE \cite{chawla2002smote}, a method that produces artificial minority samples by interpolating neighboring minority class instances. Several variants of SMOTE \cite{maciejewski2011local, he2008adasyn} improve upon the original algorithm by also taking majority class neighbors into consideration. Algorithmic methods adjust the learning or decision process in a way that directly imposes a heavier cost on misclassifying the minority class while avoiding the error-prone issue in broadened decision regions.\\

\section{DAI ALGORITHM}

Here we define $A$ as the label matrix with size $m\times n$, $m$ equals to the number of training samples and $n$ equals to the number of attributes each image has. To be specific, PA-100K dataset has 100,000 pedestrian images for training and each image has 26 binary attributes, so here $A$ is a binary matrix with size $100000\times26$. $r$ is the reweight vector which has the length equals to $m$, it represents the number of samples exists in the sub-balance dataset after applying DAI on the whole training dataset. $1$ in formula (\ref{p_formular}) represents the vector with all items equals to 1. After reweighting, the proportion of positive sample in each label can be expressed as formula (\ref{p_formular}). The $\circ$ represents the hadamard product:

\begin{equation}
{p} = \frac{1^{T} ((r1^{T})\circ (A))}{r^{T}1} \label{p_formular}
\end{equation}

Then we want the labels with minority positive samples to be more balanced by re-sampling. Here we give the goal to minimize as formula (\ref{goal}):

\begin{equation}
    f(r)={max}(0,(p_{ideal} - p)^{3}) \quad {for}\quad {r} \ge 0 \label{goal}
\end{equation}

The ${p}_{ideal}$ represents the ideal proportion of positive samples in each label. Empirically, we set each item of ${p}_{ideal}$ to be 0.6. $r$ is restricted to be bigger or equal than negative value is meaningless. 
To minimize the function (\ref{goal}), we change it into the unrestricted version as formula (\ref{goal_2}), we can minimize it with gradient descent, e.g. Adam \cite{kingma2014adam}:

\begin{equation}
    f(r)={max}(0,(p_{ideal} - p)^{3}) - \lambda {r} \quad \lambda > 0
    \label{goal_2}
\end{equation}

$\lambda$ equals to the regularization parameter. After computing the $r$, we rescale it by multiplying a constant e.g. 5 or 10, then get the integer of it. The integer represents the number of each sample exists in the sub-balance dataset.

Figure \ref{fig_34} demonstrates the distribution of each label before and after DAI applies. We can see that the positive proportion of each label gets more balanced.

\begin{figure}[!ht]
\setlength{\abovecaptionskip}{0.cm}
	\begin{center}
		\subfigure[]{
			\label{fig_1}
			\centering
			\includegraphics[width=2.5 in]{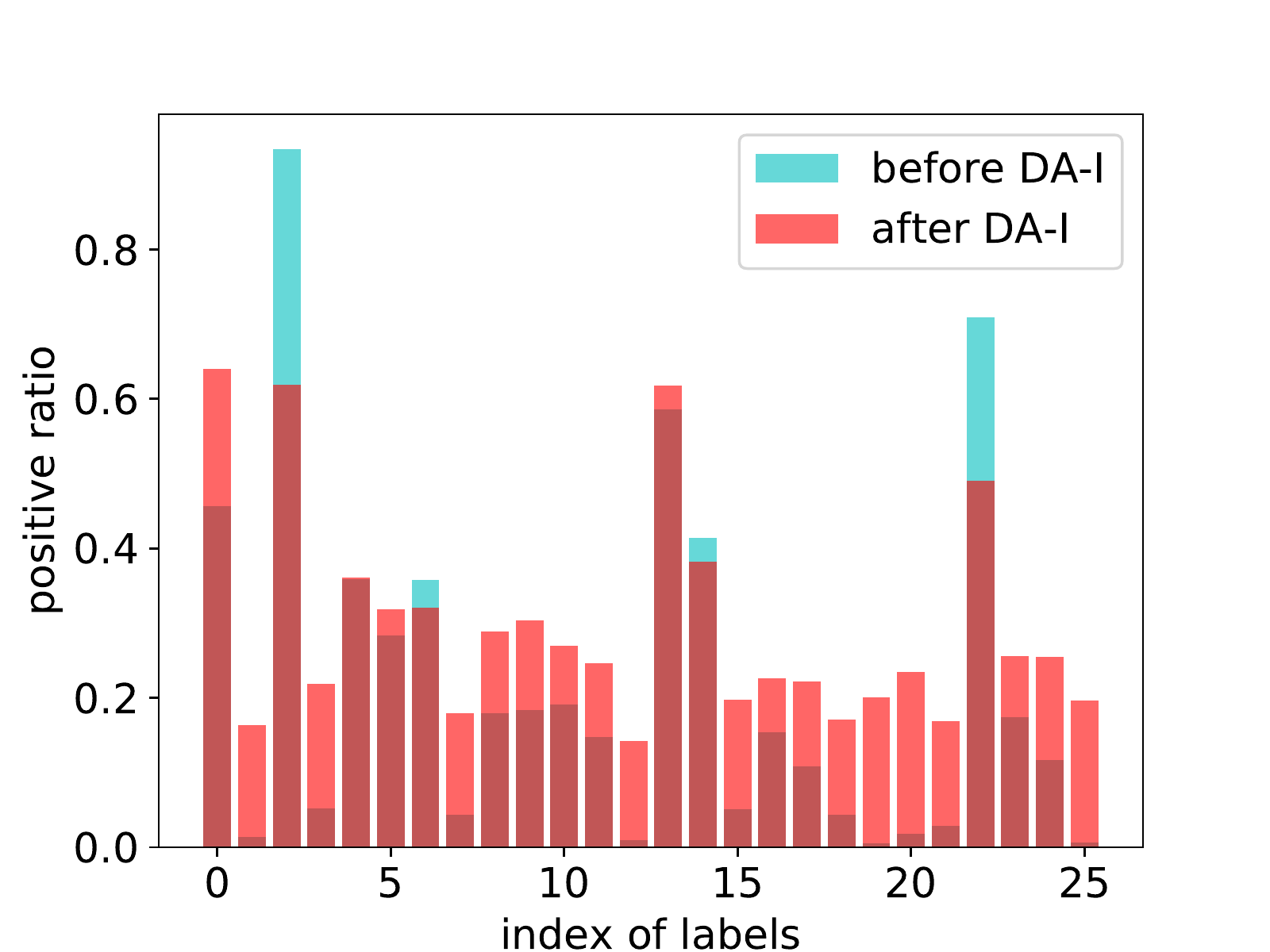}      
		}\hspace{3mm}
		\subfigure[]{
			\label{fig_2}
			\centering
			\includegraphics[width=2.5 in]{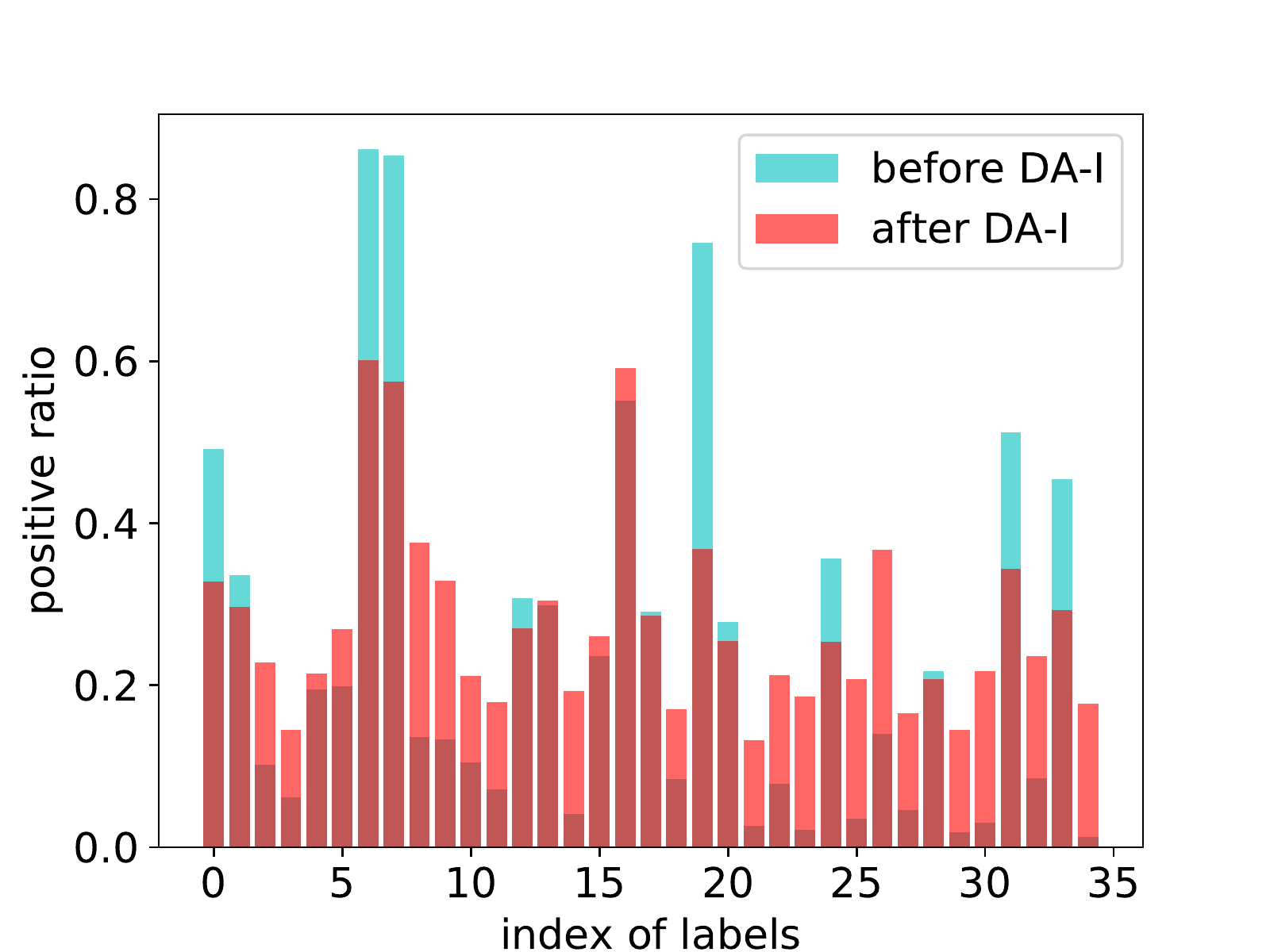}     
		}\hspace{-30mm}
	\end{center}
	\caption{Two figures demonstrate the distribution of positive samples in each label before and after DAI applies.} \label{fig_34}
\end{figure}

\begin{figure}[!ht]
\setlength{\abovecaptionskip}{0.cm}
	\begin{center}
		\subfigure[]{
			\label{fig_1}
			\centering
			\includegraphics[width=0.5 in]{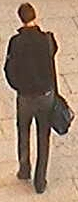}      
		}\hspace{3mm}
	    \subfigure[]{
			\label{fig_1}
			\centering
			\includegraphics[width=2 in]{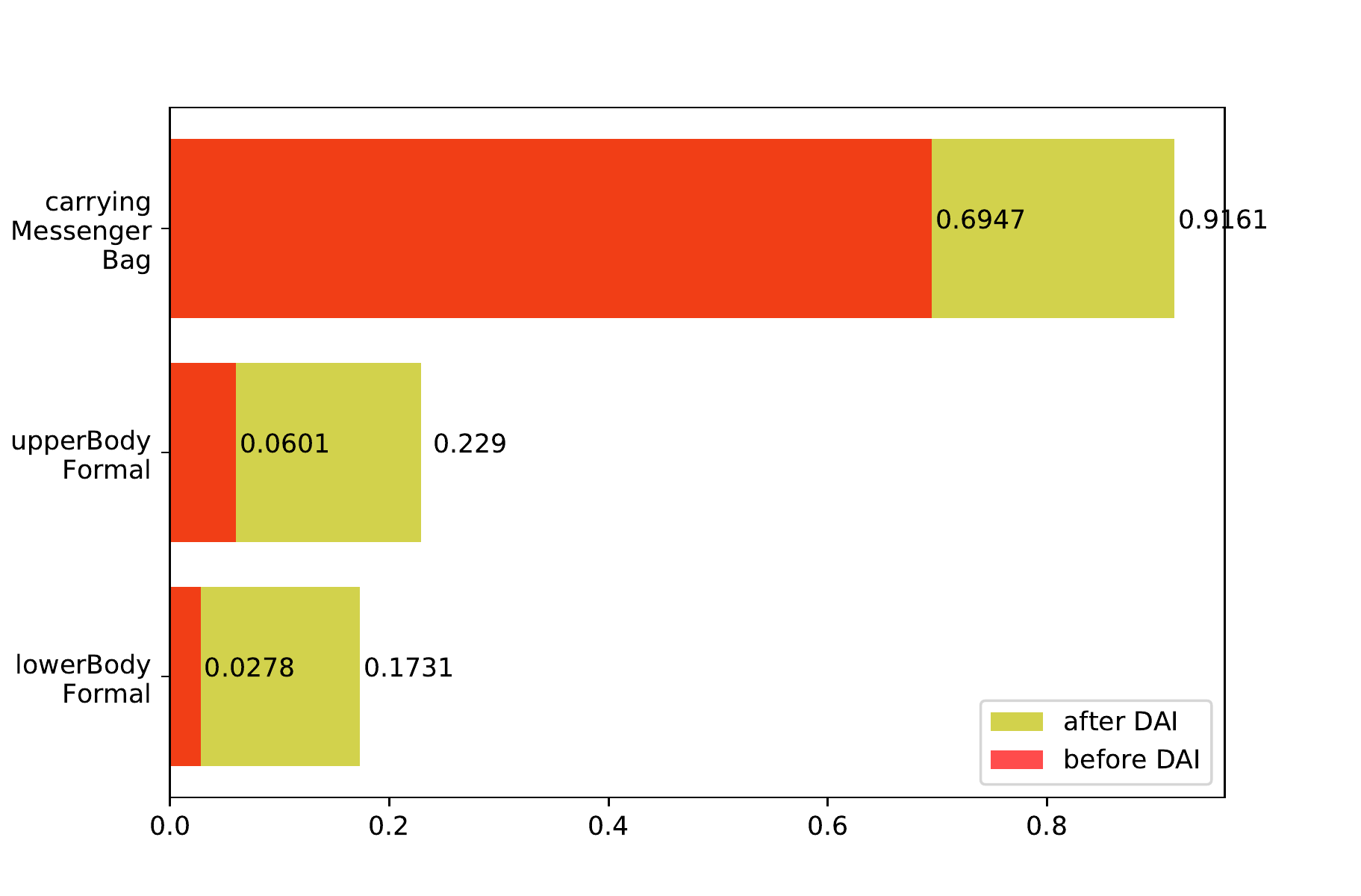}      
		}\hspace{-30mm}
	\end{center}
	\caption{The prediction results of one pedestrian. (a) one person in PETA dataset and (b) three attributes' forecast results before and after DAI. The attribute of lowerBodyFormal's positive ratio approximately is 0.14, upperBodyFormal is 0.13, carryingMessengerBag is 0.29.} \label{fig_34}
\end{figure}

After getting the balanced dataset(sub-balance dataset), we train the network with sub-balance dataset and the whole dataset in turn. Figure 3 shows the forecast results of one person in PETA dataset. The three attributes have fewer positive samples and after DAI, the discriminative ability improve much more.




\section{Experiments}
\label{sec:format}
\subsection{Benchmark Data}
\label{ssec:subhead}
We evalute our DAI model on three public datasets. The \textbf{PA-100K dataset} \cite{liu2017hydraplus} is a large pedestrian attribute dataset with a totally 100,000 images collected from outdoor surveillance cameras. The whole dataset is randomly split into training, validation and test sets with a ratio of 8:1:1 and is labelled by 26 binary attributes. The images are blurry due to the relatively low resolution and the positive ratio of each binary attribute is imbalanced. The \textbf{PETA dataset} \cite{deng2014pedestrian} consists of 19,000 cropped images collected from 10 small-scale surveillance datasets with the separation of 9500 training images, 1900 validation images, and 7600 test images. Each image is labeled with 61 binary and 5 multi-value attributes. Following the established protocol, we select 35 attributes for which the positive ratio is higher than 5$\%$. In order to get a strong proof of the performance of our algorithm, we also evaluate on the \textbf{RAP dataset} \cite{li2016richly}. The Richly Annotated Pedestrian (RAP) dataset contains 41,585 images gathered from indoor surveillance cameras. Each image is annotated with 72 attributes. while only 51 binary attributes with a positive ratio above 1$\%$ are used for evaluation. We split the dataset randomly into 33,268 images for training and 8,317 for testing. However, we find that the positive ratio among labels in all datasets show seriously imbalanced.

Following the settings of previous arts, We use five metrics to evaluate DAI model. For a label-based evaluation, we compute the mean of the accuracy among positive and negative examples respectively of an attribute as the mean accuracy (mA). This metrics is not affected by class imbalances and more frequent label value equally strongly than the infrequent one. However, this metric does not make sense for the consistency among attributes for a given person sample. Therefore, we further use sample-based metrics. For this way, we apply the well-known metrics including accuracy, precision, recall and F1 score on test data. A more detailed description of metrics is introduced in \cite{li2016richly}.\\

\subsection{Ablation Study}
\label{ssec:subhead}

\begin{table}[t]
\begin{tabular}{llllll}
    \toprule
    Methods &mA &F1 &Recall &Prec &Accu\\
    \midrule
    Baseline &76.88 & 87.63 & 85.23 & 90.12 & 79.56\\
    Attention \cite{hu2018squeeze} & 76.93 & 87.64 & 85.14 & 90.29 & 79.58\\
    mixup \cite{zhang2017mixup} & 60.45 & 73.27 & 59.01 & \textbf{96.60} & 58.18\\
    weighted loss & 77.04 & 87.09 & 85.04 & 89.23 & 79.34\\
    focal loss & 68.95 & 83.32 & 74.53 & 94.47 & 72.21\\
    DAI &\textbf{77.89} &\textbf{87.75} &\textbf{85.37} & 90.26 & \textbf{79.71}\\
\bottomrule
\end{tabular}
\caption{Ablation Study on PA-100K: Effects of Components using DenseNet-161 as a light-weight backbone architecture.}
\end{table}

\begin{table}[t]
\begin{tabular}{llllll}   
    \toprule
    Methods &mA &F1 &Recall &Prec &Accu\\
    \midrule
    Baseline &71.59 & 86.01 & 84.17 & 88.04 & 78.24\\
    Attention \cite{hu2018squeeze} & 81.02 & 85.81 & 83.23 & 88.54 & 77.85\\
    mixup \cite{zhang2017mixup} & 81.01 & 86.26 & 83.36 & 89.37 & 78.39\\
    weighted loss & 71.29 & 85.94 & 84.23 & 87.81 & 78.15\\
    focal loss & 71.80 & 85.02 & 80.04 & \textbf{91.34} & 76.74\\
    DAI & \textbf{88.24} & \textbf{86.70} & \textbf{84.70} & 88.79 & \textbf{79.14}\\

\bottomrule
\end{tabular}
\caption{Ablation Study on PETA: Effects of Components using ResNeXt101$\_$32x8d as a light-weight backbone architecture.}
\end{table}

\begin{table}[t]
\begin{tabular}{llllll}
    \toprule
    Methods &mA &F1 &Recall &Prec &Accu\\
    \midrule
    Baseline &67.14 &74.57  & 73.56 &81.34 &63.35\\
    Attention \cite{hu2018squeeze} &74.24 &74.15 &73.48 & 85.17 &64.59\\
    mixup \cite{zhang2017mixup} & 73.68 & 72.79 & 74.16 & 86.79 & 65.49\\
    weighted loss & 70.37 &71.69 &75.60 & 82.57 & 64.59\\
    focal loss & 69.16 & 74.90 &78.43 &\textbf{87.18} &62.79\\
    DAI &\textbf{75.09} &\textbf{76.46} &\textbf{79.16} &84.27 & \textbf{66.90}\\
\bottomrule
\end{tabular}
\caption{Ablation Study on RAP: Effects of Components using ResNeXt101$\_$32x8d as a light-weight backbone architecture.}
\end{table}   

\begin{figure*}[!ht]
\setlength{\abovecaptionskip}{0.cm}
	\begin{center}
			\label{fig_1}
			\centering
			\includegraphics[width=7 in]{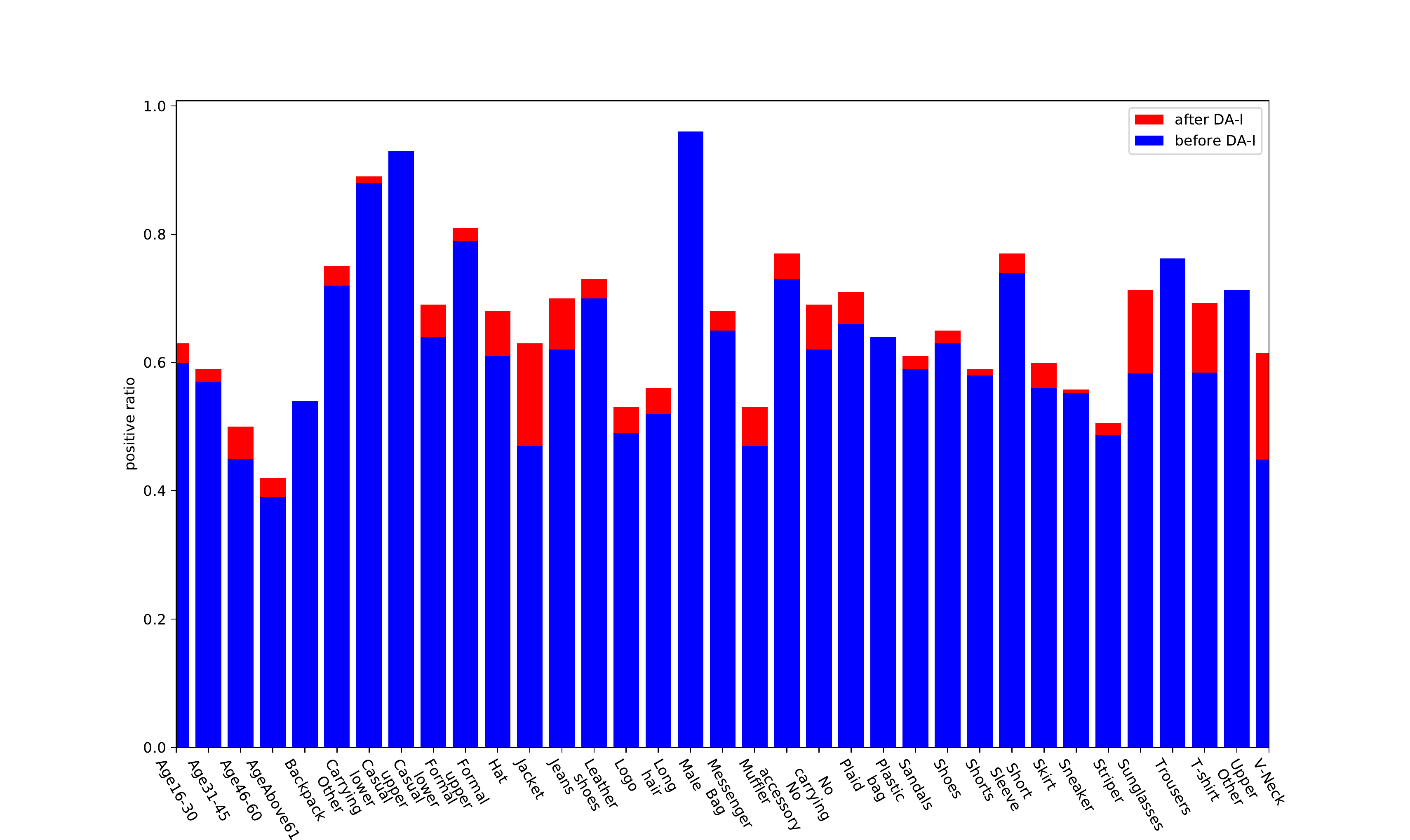}      
	\end{center}
	\caption{Mean accuracy scores for the whole attributes of PETA dataset by DAI and without DAI marked with red and blue bars respectively. The DAI outperforms especially on "Jacket", "Sunglasses" and "V-Neck" which have obviously increment on positive ratio after DAI.} \label{fig_34}
\end{figure*}

To better demonstrate the effectiveness and advantage of our designed model, we apply component-wise ablation studies to explicitly address the contribution of each block on PA-100K, PETA datasets and RAP dataset. After DAI, the subset of RAP has 14038 samples, PA-100K has 33982 samples, PETA has 4705 samples. The size of all subsets account for the original training set about 0.4.

We investigate to what the classical strategies for imbalanced data and the attention model affect the final performance. We observe that $1)$ Densenet with attention mechanism \cite{hu2018squeeze} performs similarly with Densenet as simple stacking the SE blocks which can't extract the masks giving emphasis to different spatial regions when fail to provide the classifier with attribute-discriminative information. $2)$ We assess how each proposed classical loss function influences the final scores. The mA of the proposed approach is 1.0 more than the primary network on PA-100K and far more than 10 percentage point on PETA. Our Densenet161 baseline achieves 76.88$\%$ mA which increases to 77.04$\%$ on PA-100K when the label weights are added while other metrics drop a little maybe because not taking the instance-level weighting into consideration. Handling class imbalance using the focal loss result in a drop in mA, F1, Recall, Accuracy compared with the primary net on PA-100K and in F1, Recall, Accuracy there is a decline on PETA. However, focal loss outperforms the other methods in sample-based precision metric on both datasets. $3)$  mixup \cite{zhang2017mixup} is a newly proposed method alleviating insensitivity to the adversarial examples which train a neural network on convex combinations of pairs of examples and labels while we use it for data augmentation. The precision reaches the best on PA-100K and RAP as well as ranks second on PETA for extremely efficient at regularizing models in computer vision. $4)$Our approach achieves competitive performance across all metrics compared with other well-known algorithm on imbalanced tasks particularly on mA. However, metrics on RAP dataset are generally not high. The RAP has 51 attributes, which is much larger than the other two sets. It's hard to mining the correlation between these labels and to balance every attribute well.

\subsection{Quantitative Comparison with Prior Arts}
\label{ssec:subhead}
\begin{table}[t]
\begin{tabular}{llllll}   
    \toprule
    Methods &F1 &Accu &Prec &Recall &mA\\
    \midrule
    LG-Net \cite{liu2018localization} & 85.04 & 75.55 & 86.99 & 83.17 & 76.96\\
    VeSPA \cite{sarfraz2017deep} & 83.02 & 73.00 & 84.99 & 81.49 & 76.32\\
    HP-net \cite{liu2017hydraplus} & 82.53 & 72.19 & 82.97 & 82.09 & 74.21\\
    DeepMAR \cite{li2015multi} & 81.32 & 70.39 & 82.24 & 80.42 & 72.70\\
    Densenet\\with DAI  & \textbf{87.75} & \textbf{79.71} & \textbf{90.26} & \textbf{85.37} & \textbf{77.89}\\

\bottomrule
\end{tabular}
\caption{Evaluation results on PA-100K compared with the state-of-the-art models.}
\end{table}

\begin{table}[t]
\begin{tabular}{llllll}   
    \toprule
    Methods &F1 &Accu &Prec &Recall &mA\\
    \midrule
    GRL\cite{zhao2018grouping} & 86.51 & \quad - & 84.34 & \textbf{88.82} & 86.70\\
    VAA \cite{sarafianos2018deep}& 86.46 & 78.56 & 86.79 & 86.12 & 84.59\\
    VeSPA \cite{sarfraz2017deep} & 85.49 & 77.73 & 86.18 & 84.81 & 83.45\\
    JRL \cite{wang2017attribute} & 85.42 & \quad - & 86.03 & 85.34 & 85.67\\
    ResNeXt\\with DAI & \textbf{86.70} & \textbf{79.14} & \textbf{88.79} & 84.70 & \textbf{88.24}\\

\bottomrule
\end{tabular}
\caption{Evaluation results on PETA compared with the state-of-the-art models.}
\end{table}

\textbf{Implementation Details. }Our model is finetuned from the Densenet161 for PA-100K and ResNeXt101$\_$32x8d for PETA pretrained from ImageNet image classification task. The optimization algorithm used in training the proposed model is SGD. The initial learning rate of training is 0.1 and reduced to 1e-8 by a factor of 0.1 at last. We also use data augmentation of randomErasing, randomHorizontalFlip, and randomCrop on all experiments. Furthermore, we apply a sub-all alternately training strategy to gain a better feature extractor and classifier. All ablation experiments and comparative experiments are conducted under the same configuration except for the algorithms.

\textbf{Evaluation Results. }We compare the performance of our DAI submodel with the backbone of Densenet161 with several recent state-of-the-art pedestrian attribute recognition tasks, including LG-Net, VeSPA, HP-net, DeepMAR. Tables 3 and 4 show evaluations on PA-100K and PETA respectively. We observe that the proposed approach achieves competitive performance across all metrics and yields notable improvements over the previous state-of-the-art on all scores on PA-100K dataset while the example-based Recall is lower than that of the GRL on PETA dataset. The strong F1 values indicate a good precision-recall tradeoff of our approach. However, the label based results of mA achieve most advanced on both datasets for the jointly trained backbone. Long-term training with possible overfitting will increase the Precision metric further at the cost of example-based Recall which may results in the drop of Recall on PETA. Notably, we achieve better upon VeSPA in all evaluation metrics even though they train their view-aware attribute prediction model with additional viewpoint information. Furthermore, our model surpasses upon the HP-net which captures multiple attention multi-directionally. We believe that the example-based metrics are more relevant to real-life applications as they measure the consistency of an attribute-based description of a person which is more important for communicating such characterization to security personnel.

\section{Conclusion}
In this work, we present a unified re-sampling algorithm to transform a multi-label imbalance dataset to a more balanced sub-balance dataset. Our results show that our algorithm addresses imbalanced datasets in multi-label scenarios well especially in the field of Pedestrian attributes recognition. The main idea of integrating both oversampling and undersampling by reweighting the samples helps learn attributes that take up a small portion in a dataset very well. In comparison to the published state-of-the-art Which uses attention model or viewpoint information, our results show that explicitly increasing the proportion of less attributes is proved useful for person attribute recognition. In addition to providing convincing attribute predictions, DAI may also aid in person re-identification.

\bibliographystyle{IEEEbib}
\bibliography{refs}

\begin{thebibliography}{10}

\bibitem{deng2014pedestrian}
Yubin Deng, Ping Luo, Chen~Change Loy, and Xiaoou Tang,
\newblock ``Pedestrian attribute recognition at far distance,''
\newblock in {\em Proceedings of the 22nd ACM international conference on
  Multimedia}. ACM, 2014, pp. 789--792.

\bibitem{li2015multi}
Dangwei Li, Xiaotang Chen, and Kaiqi Huang,
\newblock ``Multi-attribute learning for pedestrian attribute recognition in
  surveillance scenarios,''
\newblock in {\em 2015 3rd IAPR Asian Conference on Pattern Recognition
  (ACPR)}. IEEE, 2015, pp. 111--115.

\bibitem{liu2018localization}
Pengze Liu, Xihui Liu, Junjie Yan, and Jing Shao,
\newblock ``Localization guided learning for pedestrian attribute
  recognition,''
\newblock {\em arXiv preprint arXiv:1808.09102}, 2018.

\bibitem{liu2017hydraplus}
Xihui Liu, Haiyu Zhao, Maoqing Tian, Lu~Sheng, Jing Shao, Shuai Yi, Junjie Yan,
  and Xiaogang Wang,
\newblock ``Hydraplus-net: Attentive deep features for pedestrian analysis,''
\newblock in {\em Proceedings of the IEEE international conference on computer
  vision}, 2017, pp. 350--359.

\bibitem{sarfraz2017deep}
M~Saquib Sarfraz, Arne Schumann, Yan Wang, and Rainer Stiefelhagen,
\newblock ``Deep view-sensitive pedestrian attribute inference in an end-to-end
  model,''
\newblock {\em arXiv preprint arXiv:1707.06089}, 2017.

\bibitem{zhao2018grouping}
Xin Zhao, Liufang Sang, Guiguang Ding, Yuchen Guo, and Xiaoming Jin,
\newblock ``Grouping attribute recognition for pedestrian with joint recurrent
  learning.,''
\newblock in {\em IJCAI}, 2018, pp. 3177--3183.

\bibitem{wang2017attribute}
Jingya Wang, Xiatian Zhu, Shaogang Gong, and Wei Li,
\newblock ``Attribute recognition by joint recurrent learning of context and
  correlation,''
\newblock in {\em Proceedings of the IEEE International Conference on Computer
  Vision}, 2017, pp. 531--540.

\bibitem{huang2019deep}
Chen Huang, Yining Li, Change~Loy Chen, and Xiaoou Tang,
\newblock ``Deep imbalanced learning for face recognition and attribute
  prediction,''
\newblock {\em IEEE transactions on pattern analysis and machine intelligence},
  2019.

\bibitem{sarafianos2018deep}
Nikolaos Sarafianos, Xiang Xu, and Ioannis~A Kakadiaris,
\newblock ``Deep imbalanced attribute classification using visual attention
  aggregation,''
\newblock in {\em Proceedings of the European Conference on Computer Vision
  (ECCV)}, 2018, pp. 680--697.

\bibitem{ting2000comparative}
Kai~Ming Ting,
\newblock ``A comparative study of cost-sensitive boosting algorithms,''
\newblock in {\em In Proceedings of the 17th International Conference on
  Machine Learning}. Citeseer, 2000.

\bibitem{zadrozny2003cost}
Bianca Zadrozny, John Langford, and Naoki Abe,
\newblock ``Cost-sensitive learning by cost-proportionate example weighting.,''
\newblock in {\em ICDM}, 2003, vol.~3, p. 435.

\bibitem{zhou2005training}
Zhi-Hua Zhou and Xu-Ying Liu,
\newblock ``Training cost-sensitive neural networks with methods addressing the
  class imbalance problem,''
\newblock {\em IEEE Transactions on knowledge and data engineering}, vol. 18,
  no. 1, pp. 63--77, 2005.

\bibitem{chawla2002smote}
Nitesh~V Chawla, Kevin~W Bowyer, Lawrence~O Hall, and W~Philip Kegelmeyer,
\newblock ``Smote: synthetic minority over-sampling technique,''
\newblock {\em Journal of artificial intelligence research}, vol. 16, pp.
  321--357, 2002.

\bibitem{he2009learning}
Haibo He and Edwardo~A Garcia,
\newblock ``Learning from imbalanced data,''
\newblock {\em IEEE Transactions on knowledge and data engineering}, vol. 21,
  no. 9, pp. 1263--1284, 2009.

\bibitem{maciejewski2011local}
Tomasz Maciejewski and Jerzy Stefanowski,
\newblock ``Local neighbourhood extension of smote for mining imbalanced
  data,''
\newblock in {\em 2011 IEEE Symposium on Computational Intelligence and Data
  Mining (CIDM)}. IEEE, 2011, pp. 104--111.

\bibitem{layne2014attributes}
Ryan Layne, Timothy~M Hospedales, and Shaogang Gong,
\newblock ``Attributes-based re-identification,''
\newblock in {\em Person Re-Identification}, pp. 93--117. Springer, 2014.

\bibitem{layne2012person}
Ryan Layne, Timothy~M Hospedales, Shaogang Gong, and Q~Mary,
\newblock ``Person re-identification by attributes.,''
\newblock in {\em Bmvc}, 2012, vol.~2, p.~8.

\bibitem{he2008adasyn}
Haibo He, Yang Bai, Edwardo~A Garcia, and Shutao Li,
\newblock ``Adasyn: Adaptive synthetic sampling approach for imbalanced
  learning,''
\newblock in {\em 2008 IEEE International Joint Conference on Neural Networks
  (IEEE World Congress on Computational Intelligence)}. IEEE, 2008, pp.
  1322--1328.

\bibitem{kingma2014adam}
Diederik~P Kingma and Jimmy Ba,
\newblock ``Adam: A method for stochastic optimization,''
\newblock {\em arXiv preprint arXiv:1412.6980}, 2014.

\bibitem{li2016richly}
Dangwei Li, Zhang Zhang, Xiaotang Chen, Haibin Ling, and Kaiqi Huang,
\newblock ``A richly annotated dataset for pedestrian attribute recognition,''
\newblock {\em arXiv preprint arXiv:1603.07054}, 2016.

\bibitem{hu2018squeeze}
Jie Hu, Li~Shen, and Gang Sun,
\newblock ``Squeeze-and-excitation networks,''
\newblock in {\em Proceedings of the IEEE conference on computer vision and
  pattern recognition}, 2018, pp. 7132--7141.

\bibitem{zhang2017mixup}
Hongyi Zhang, Moustapha Cisse, Yann~N Dauphin, and David Lopez-Paz,
\newblock ``mixup: Beyond empirical risk minimization,''
\newblock {\em arXiv preprint arXiv:1710.09412}, 2017.

\end{thebibliography}

\end{document}